\documentclass[a4paper,11pt]{article}

\usepackage{booktabs} 

\usepackage[ruled]{algorithm2e} 

\SetAlFnt{\small}
\SetAlCapFnt{\small}
\SetAlCapNameFnt{\small}
\SetAlCapHSkip{0pt}
\IncMargin{-\parindent}

\usepackage{graphicx}
\graphicspath{ {figure/} }
\usepackage{multirow}
\usepackage{enumitem}
\usepackage{ textcomp }
\usepackage{xspace}
\usepackage{parskip}
\makeatletter
\DeclareRobustCommand\onedot{\futurelet\@let@token\@onedot}
\def\@onedot{\ifx\@let@token.\else.\null\fi\xspace}

\def\etc{\emph{etc}\onedot}

\makeatother

\newcommand{\rom}[1]{\expandafter{\romannumeral #1\relax}}

\usepackage{fourier}
\usepackage{color}
 \usepackage{graphicx}
\usepackage{url}
\usepackage[affil-it]{authblk}
\usepackage{amsmath}
\usepackage{wrapfig}
\usepackage{xspace}

\usepackage[T1]{fontenc}
\usepackage{times}
\usepackage{wrapfig}
\usepackage{titlesec}
\titlespacing\section{0pt}{12pt plus 4pt minus 2pt}{0pt plus 2pt minus 2pt}
\titlespacing\subsection{0pt}{12pt plus 4pt minus 2pt}{0pt plus 2pt minus 2pt}
\titlespacing\subsubsection{0pt}{12pt plus 4pt minus 2pt}{0pt plus 2pt minus 2pt}
\begin{document}

\title{A Geometry-Sensitive Approach for Photographic Style Classification}

\author[1]{Koustav Ghosal}
\author[1,2]{Mukta Prasad}
\author[1]{Aljosa Smolic}
\affil[1]{V-SENSE, School of Computer Science and Statistics, Trinity College Dublin}
\affil[2]{Daedalean, Zurich}
\date{}
\maketitle
\thispagestyle{empty}
\begin{abstract}
\noindent Photographs are characterized by different compositional attributes like the Rule of Thirds, depth of field, vanishing-lines \etc. The presence or absence of one or more of these attributes contributes to the overall artistic value of an image. 
In this work, we analyze the ability of deep learning based methods to learn such photographic style attributes.
We observe that although a standard \textsc{CNN} learns the texture and appearance based features reasonably well, its understanding of global and geometric features is limited by two factors. First, the data-augmentation strategies~(cropping, warping, \etc) distort the composition of a photograph and affect the performance. Secondly, the \textsc{CNN} features, in principle, are translation-invariant and appearance-dependent. But some geometric properties important for aesthetics,~\textit{e.g.~}\textit{the Rule of Thirds}~(RoT), are position-dependent and appearance-invariant.  Therefore, we propose a novel input representation which is geometry-sensitive, position-cognizant and appearance-invariant. We further introduce a two-column \textsc{CNN} architecture that performs better than the state-of-the-art~(SoA) in photographic style classification. From our results, we observe that the proposed network learns both the geometric and appearance-based attributes better than the SoA. 
\end{abstract}
\textbf{Keywords:} Deep Learning, Convolutional Neural Networks, Computational Aesthetics
\section{Introduction}
\label{sec:intro}
\begin{wrapfigure}[21]{L}{0.5\textwidth}
\vspace{-12pt}
\begin{center}
\resizebox{.5\textwidth}{!}{
\setlength{\fboxsep}{.75 em}
\fbox{
\def\arraystretch{1.5}
\begin{tabular}{cccc}
\includegraphics[height=42pt, width = 56pt]{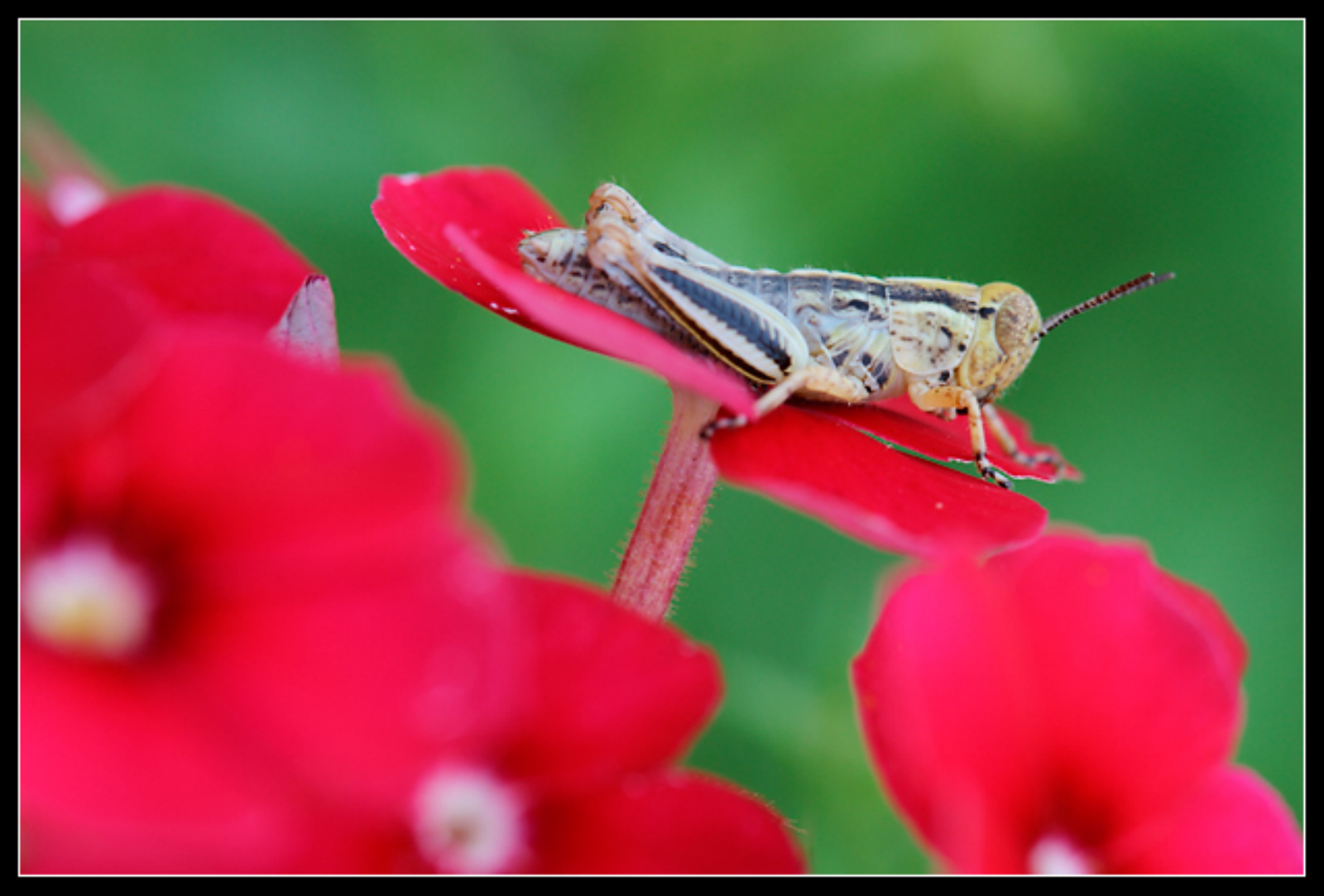}&
\includegraphics[height=42pt, width = 56pt]{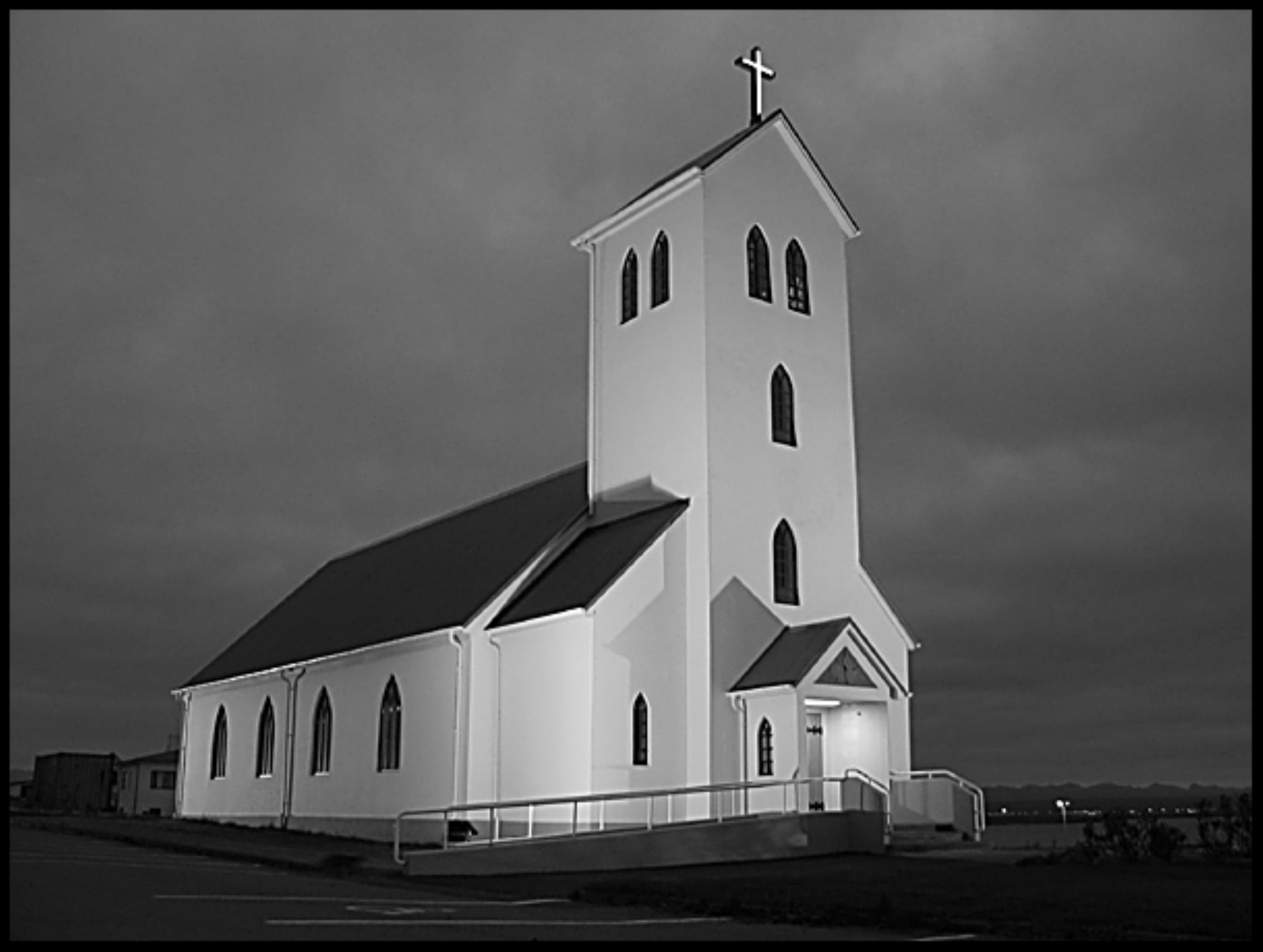}&
\includegraphics[height=42pt, width = 56pt]{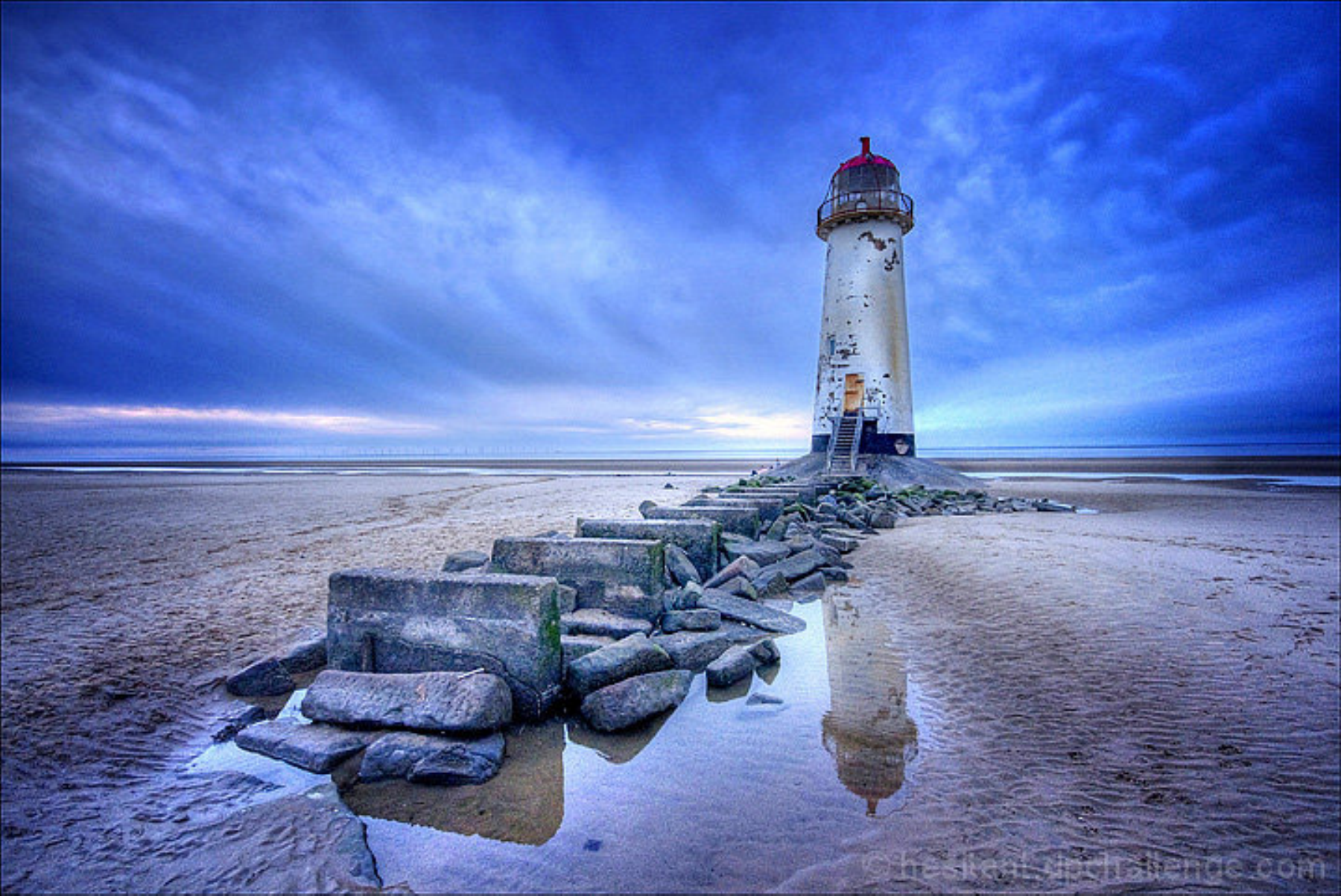}&
\includegraphics[height=42pt, width = 56pt]{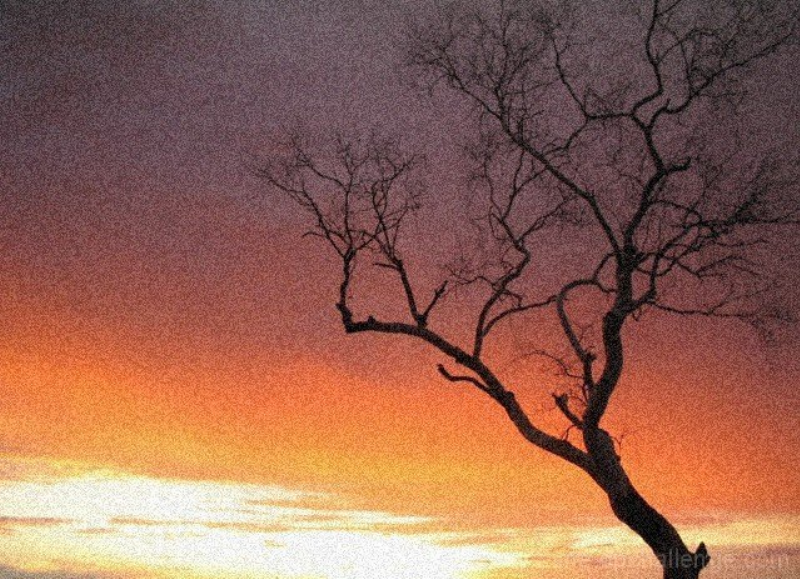}\\
\includegraphics[height=42pt, width = 56pt]{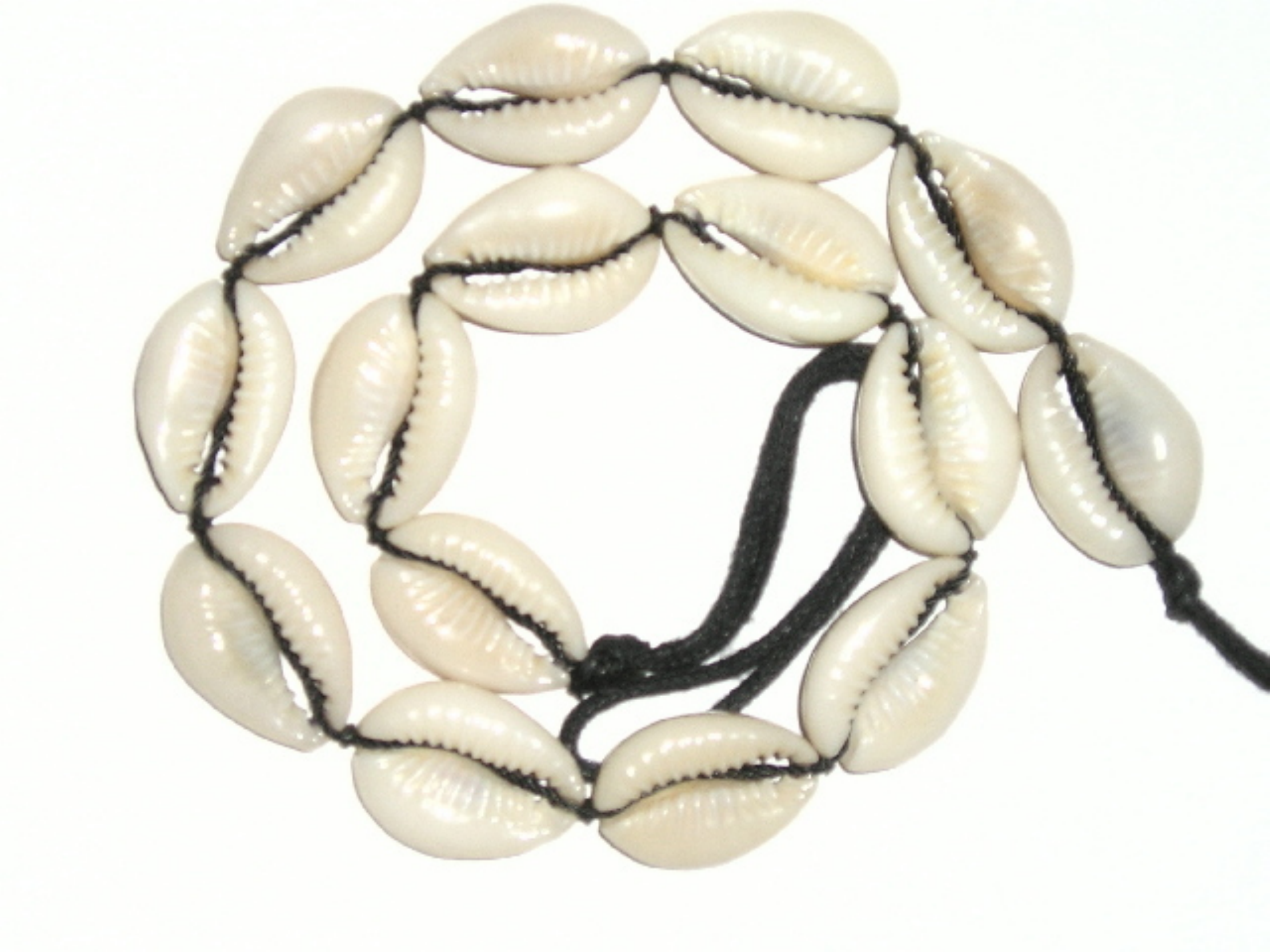}&
\includegraphics[height=42pt, width = 56pt]{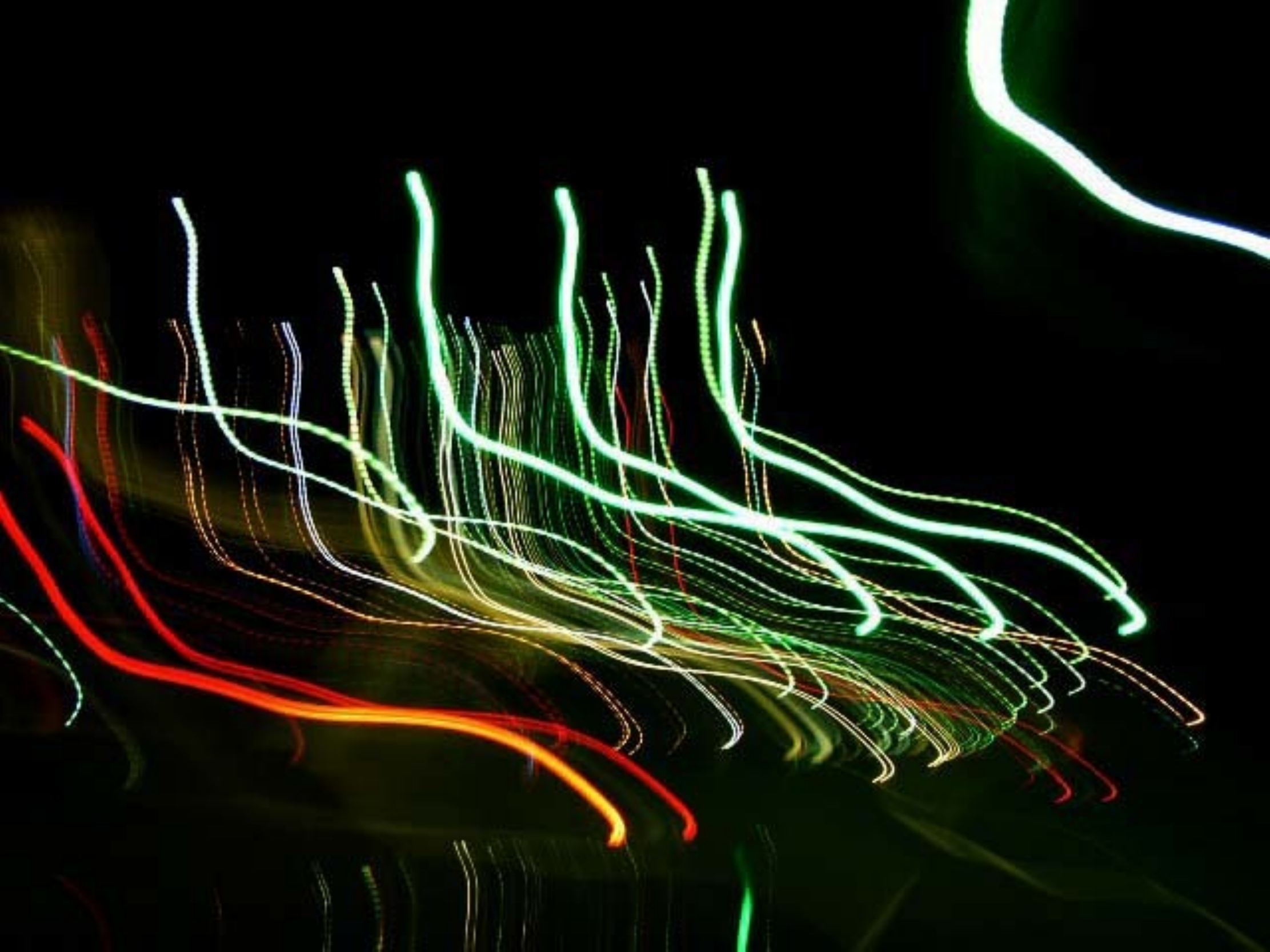}&
\includegraphics[height=42pt, width = 56pt]{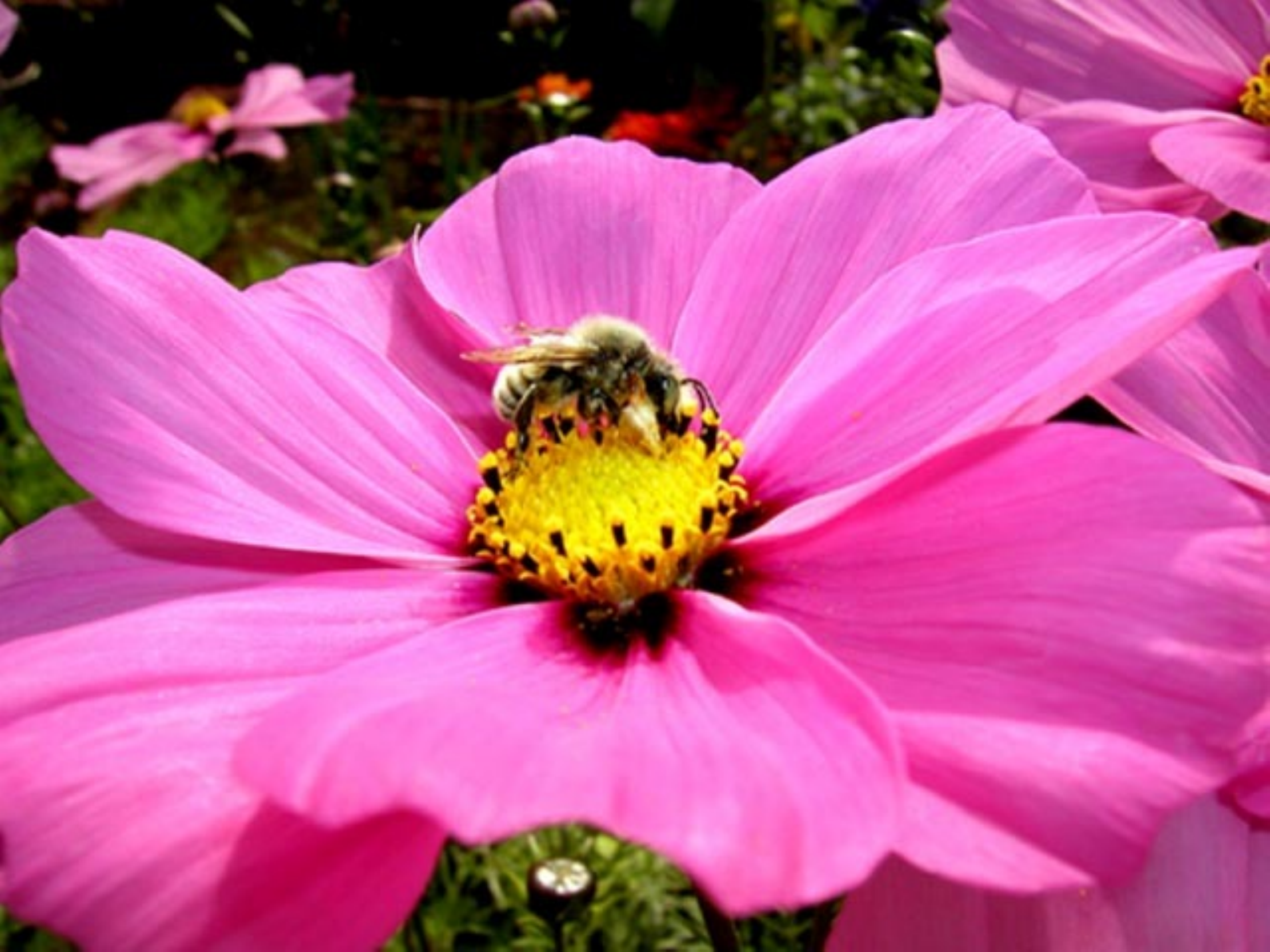}&
\includegraphics[height=42pt, width = 56pt]{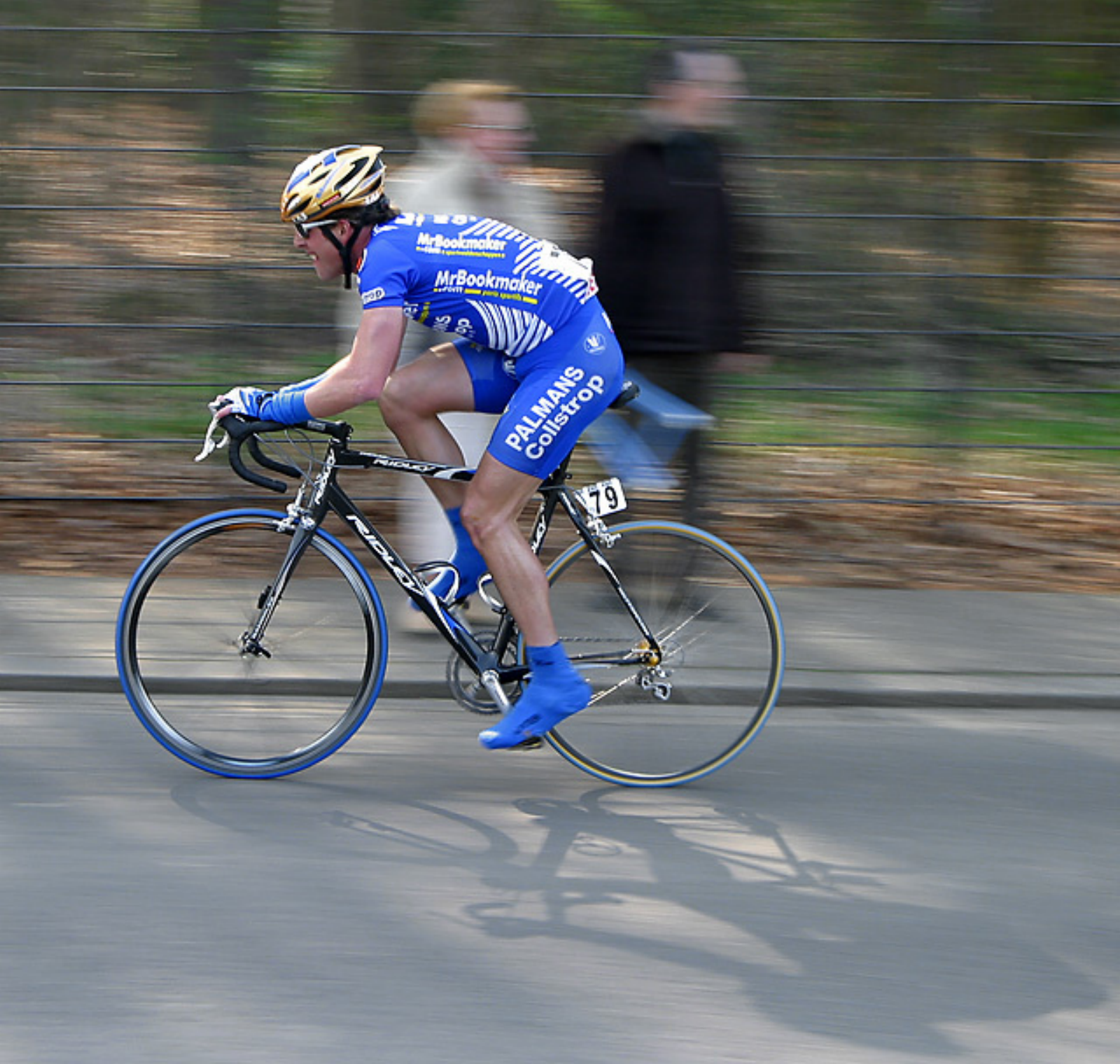}\\
\includegraphics[height=42pt, width = 56pt]{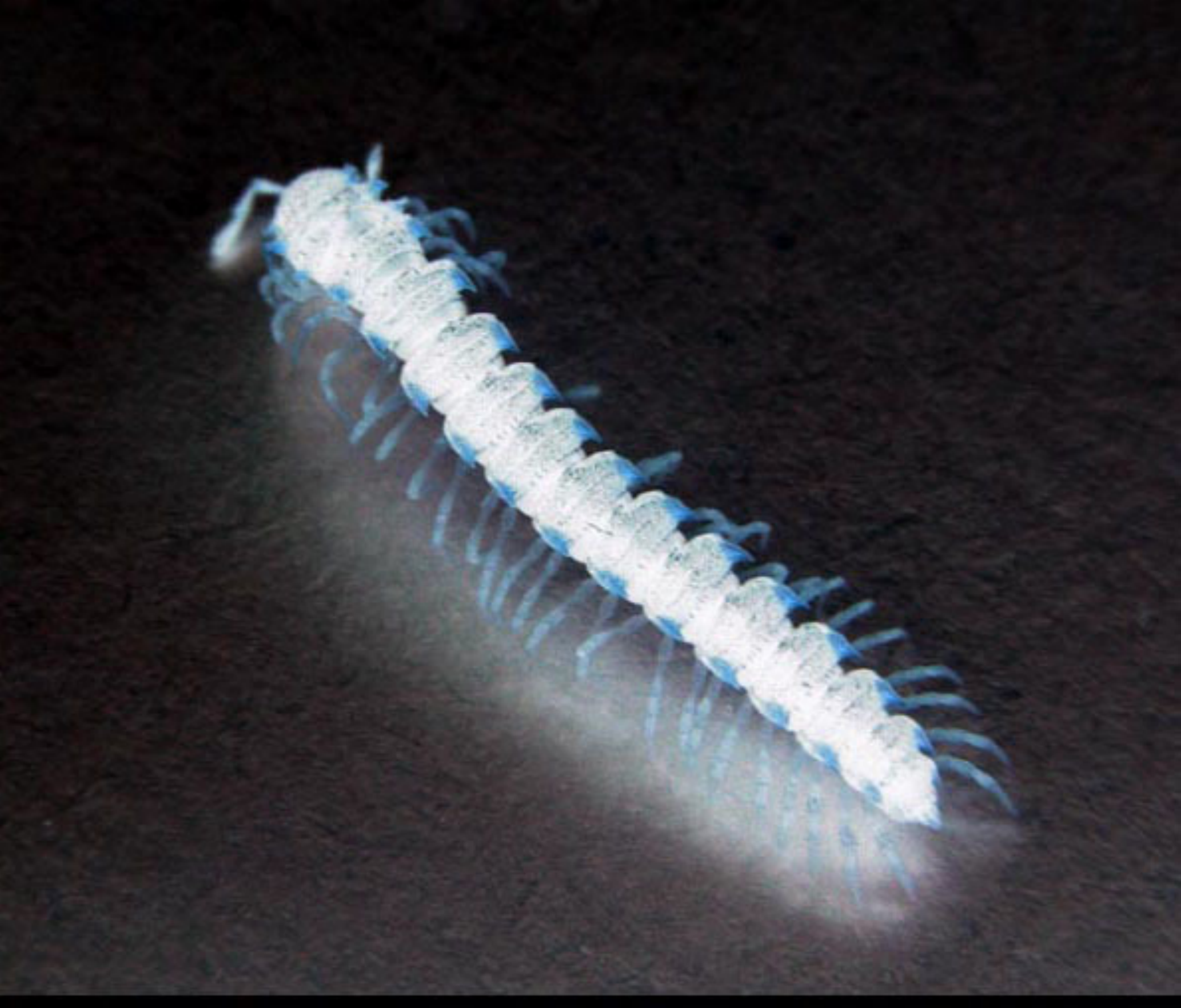}&
\includegraphics[height=42pt, width = 56pt]{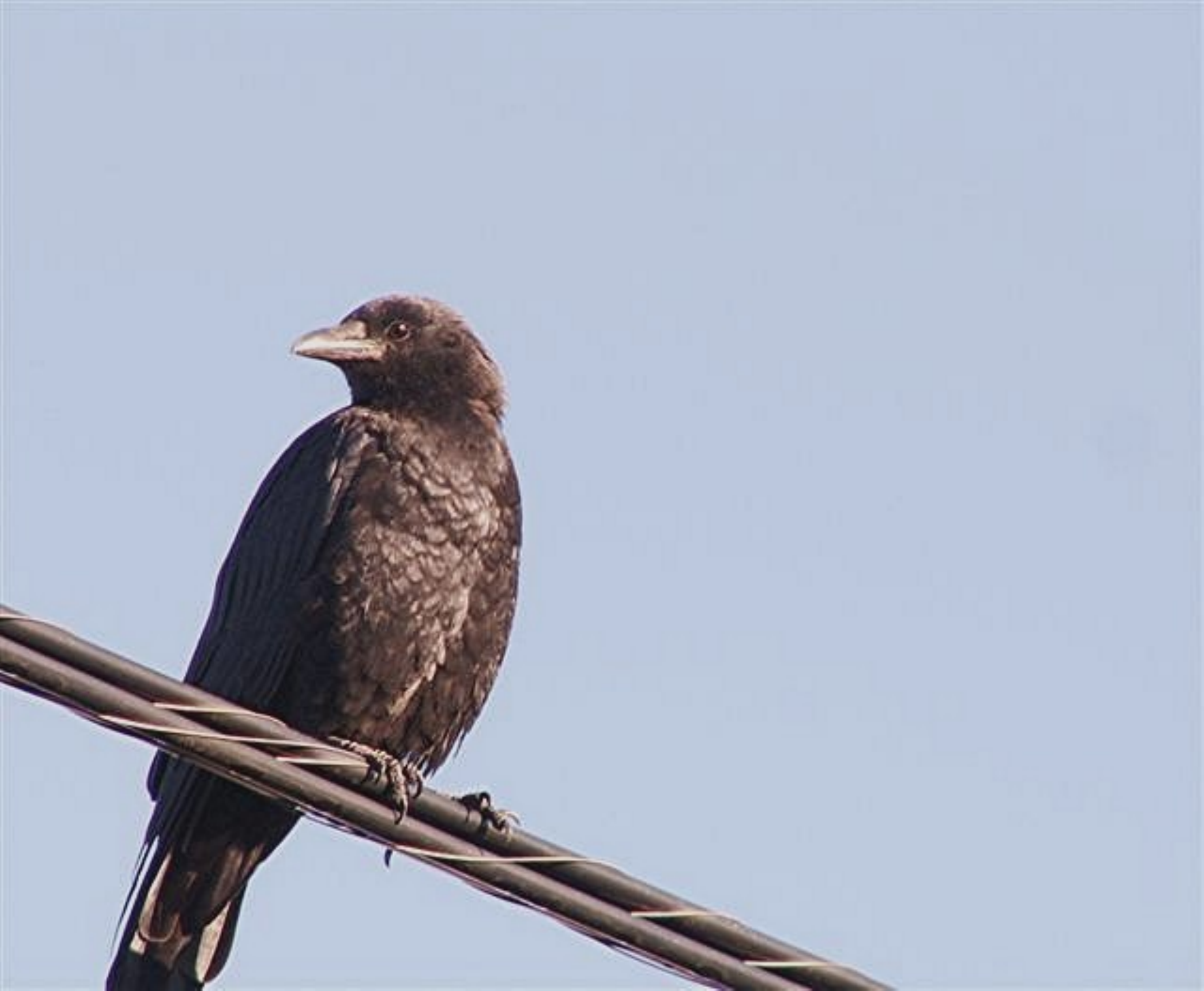}&
\includegraphics[height=42pt, width = 56pt]{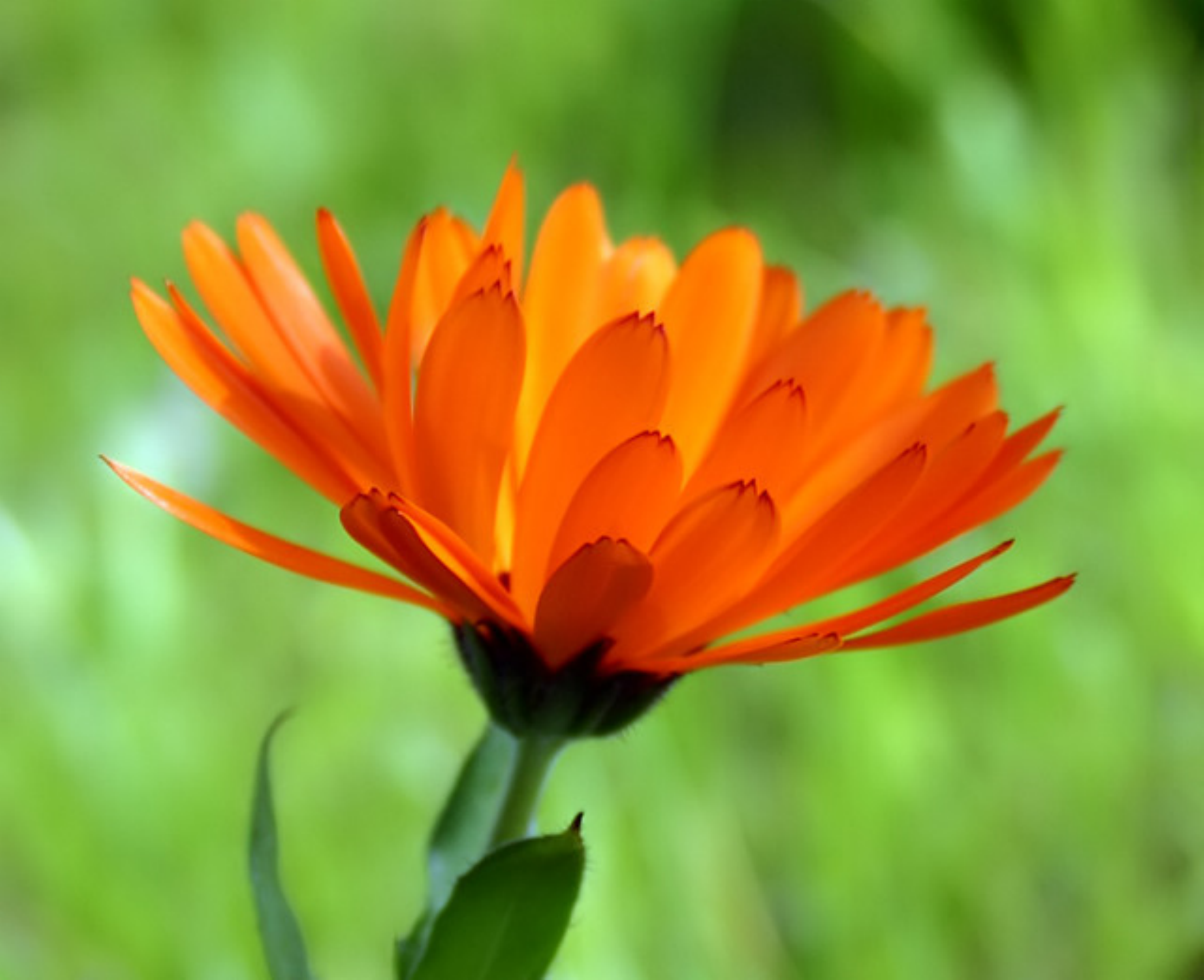}&
\includegraphics[height=42pt, width = 56pt]{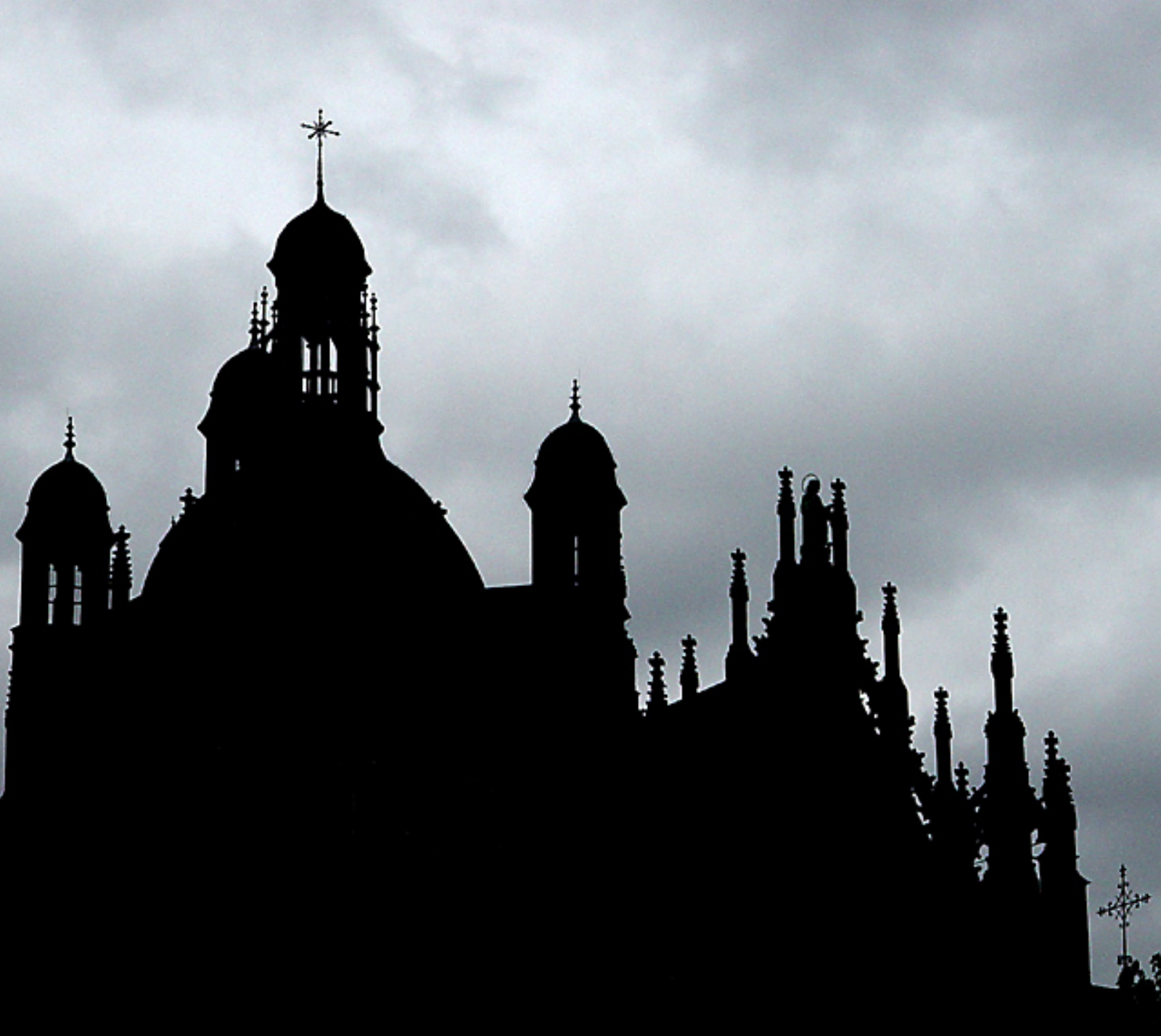}\\
\includegraphics[height=42pt, width = 56pt]{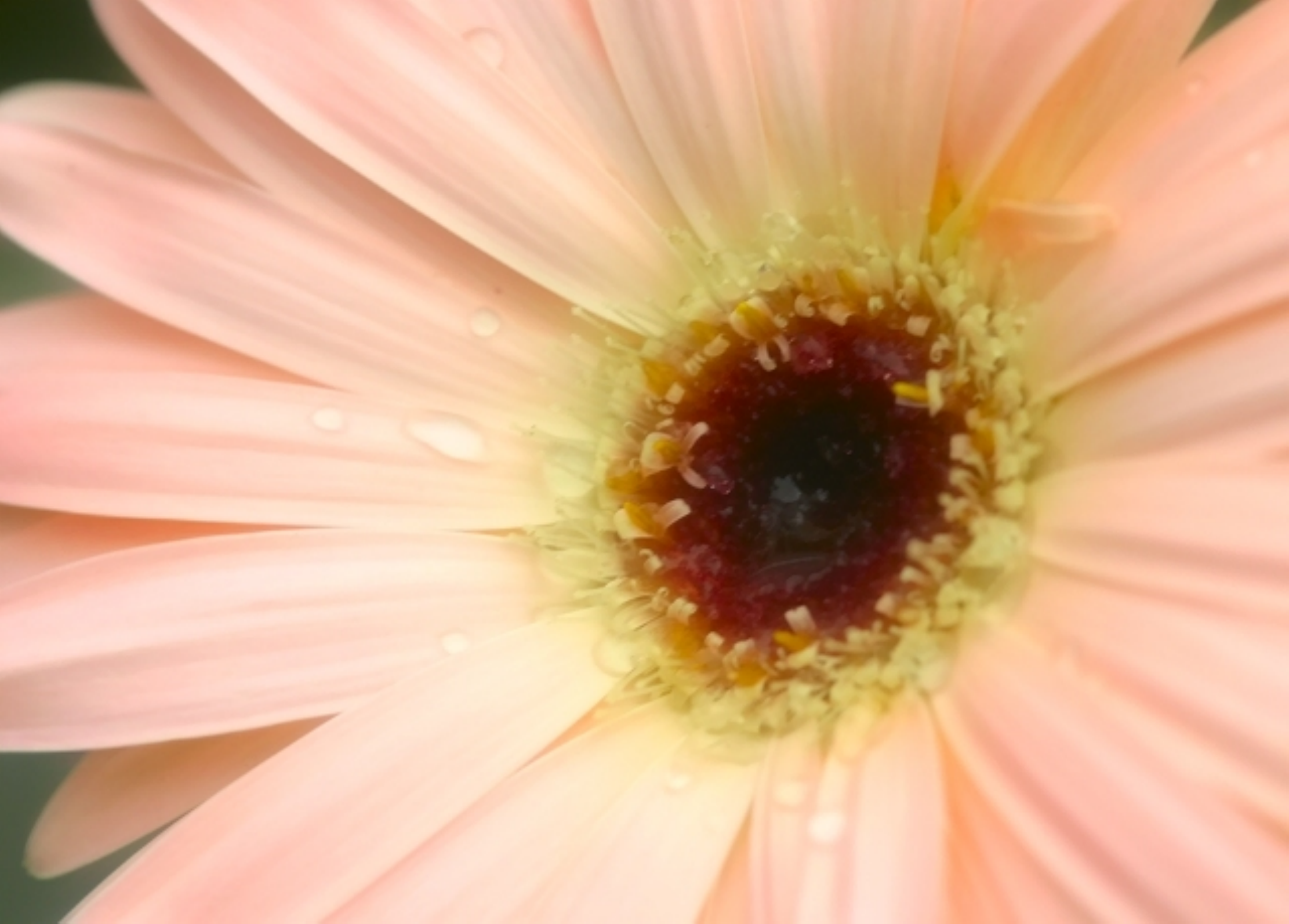}&
\includegraphics[height=42pt, width = 56pt]{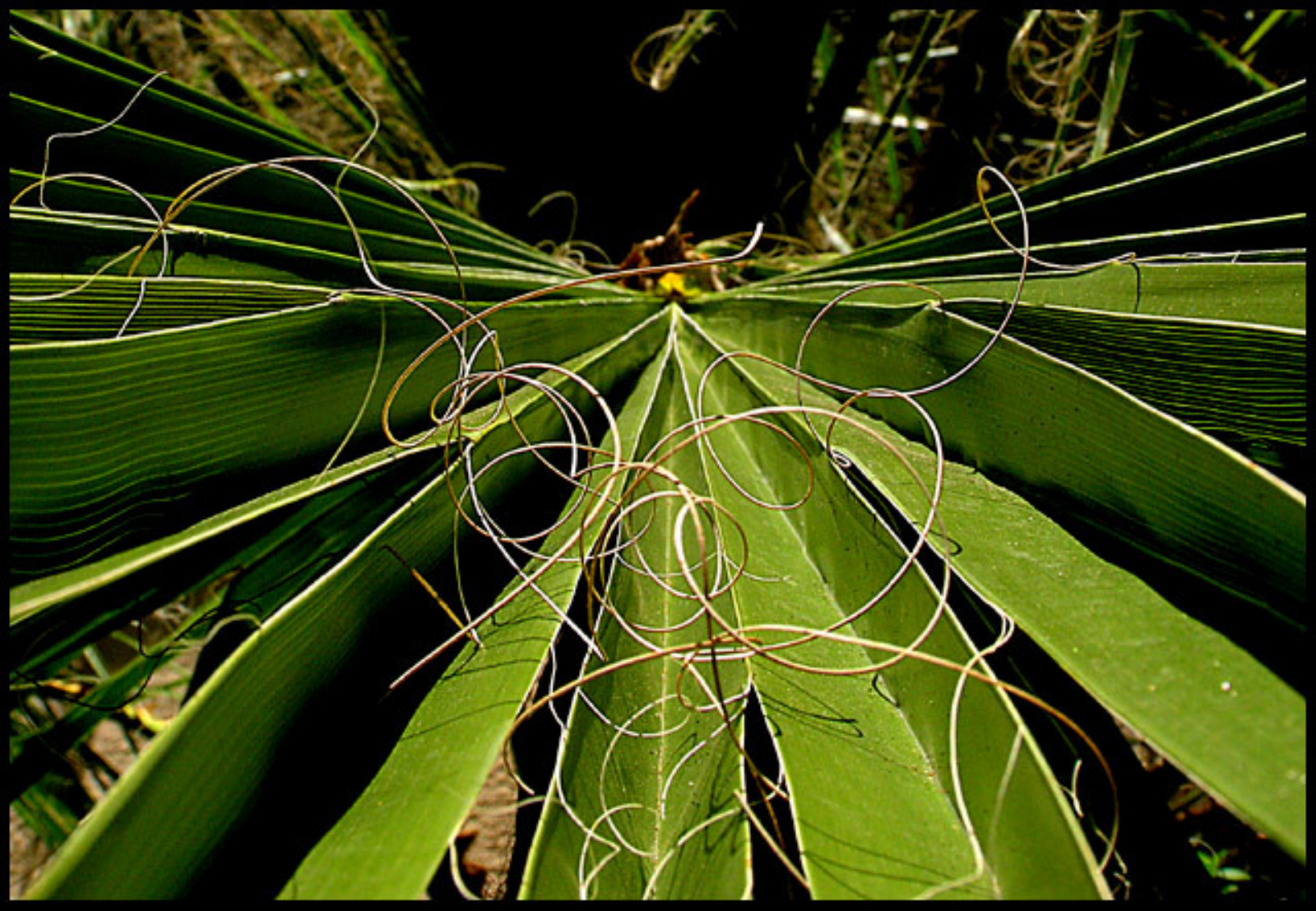}&&
\end{tabular}}}
\caption{\textbf{Example images from the AVA dataset corresponding to 14 different styles:} (L-R)
\textbf{Row 1 : }  Complementary Colors, Duotones, HDR, Image Grain. \textbf {Row 2 : } Light On White, Long Exposure, Macro, Motion Blur. \textbf{ Row 3 : } Negative Image, Rule of Thirds, Shallow DOF, Silhouettes. \textbf{ Row 4 : } Soft Focus, Vanishing Point}
\label{fig:ExampleImages}
\end{center}
\vspace{-1pt}
\end{wrapfigure}

\noindent Analyzing compositional attributes or styles is crucial for understanding the aesthetic
value of photographs. At first, the computer vision community focused on modelling the  physical properties of generic images for the 
\begin{wrapfigure}[18]{r}{0.51\textwidth}
\vspace{-10pt}
  \begin{center}
  \setlength{\fboxrule}{.125 em}
    \fbox{\includegraphics[width=0.475\textwidth]{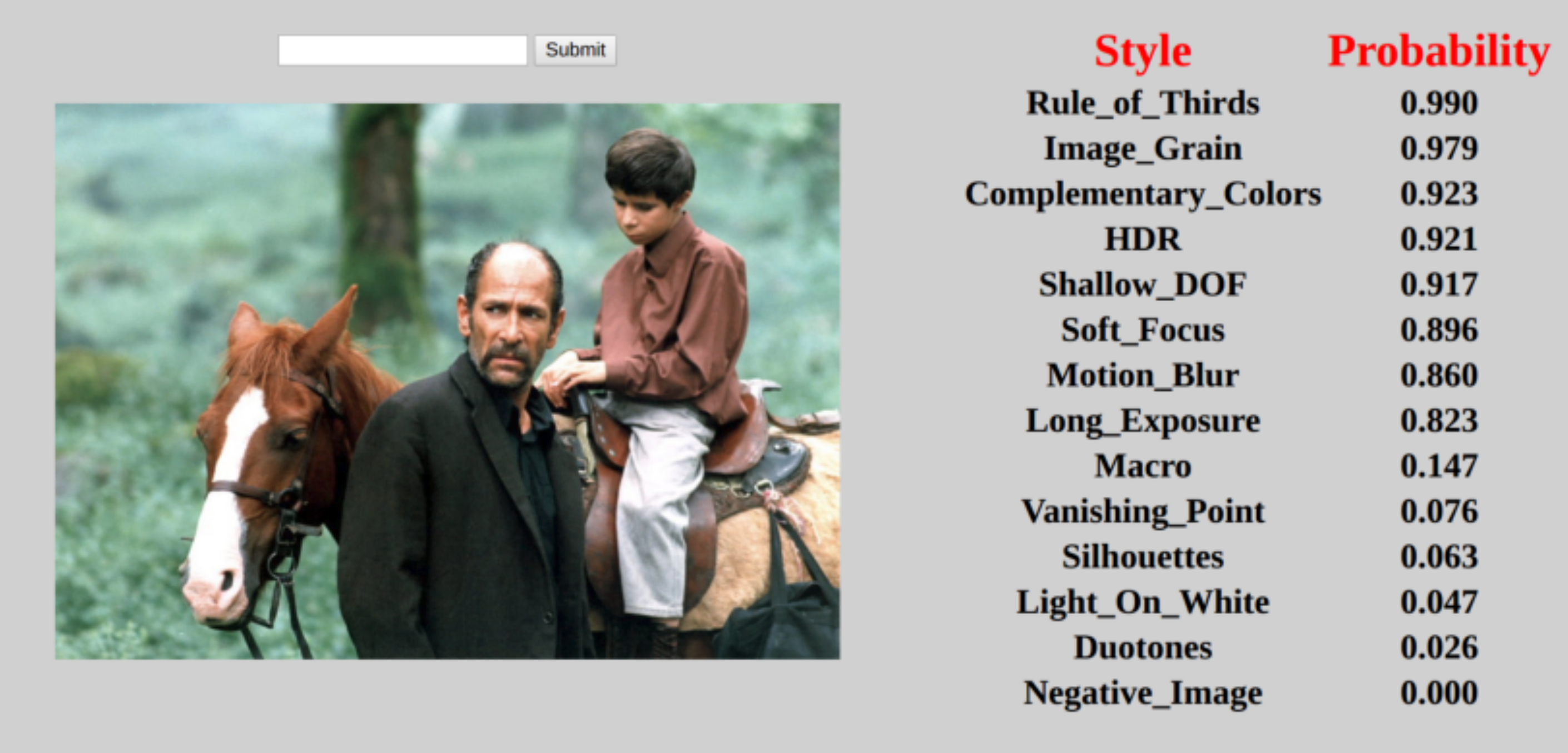}}
  \end{center}
  \vspace{-5pt}
  \caption{\textbf{Output from our network} : A screenshot from our web-based application. Attributes are shown with their probability values, ordered in descending order. This is a shot from Majid Majidi's film 'The Colours of Paradise'. We see that rule of thirds (for child's position), shallow depth of field, complementary colours (green background and reddish foreground), image grain (because of the poor video quality) are all well identified.}
  \label{fig:app}
\end{wrapfigure}
more tangible, but very hard problems of object detection, localization, segmentation, tracking,~\etc.
Popular datasets like Caltech, Pascal and ImageNet were created for training and evaluating such techniques effectively.
The maturation of recognition and scene understanding has resulted in greater interest in the analysis of the subtler, aesthetic based aspects of image understanding. Furthermore, curated datasets such as \textsc{AVA} and Flickr-Style~\cite{Karayev2014,murray2012ava} are now available and it is observed that learning from the matured areas of recognition/classification/detection transfers effectively to aesthetics and style analysis as well.\\ 
\begin{figure}
\begin{center}
\resizebox{.825\textwidth}{!}{

\begin{tabular}{cc}
\resizebox{.345\textwidth}{!}{
\setlength{\fboxrule}{.125 em}
\fbox{
\includegraphics[width=0.5\textwidth]{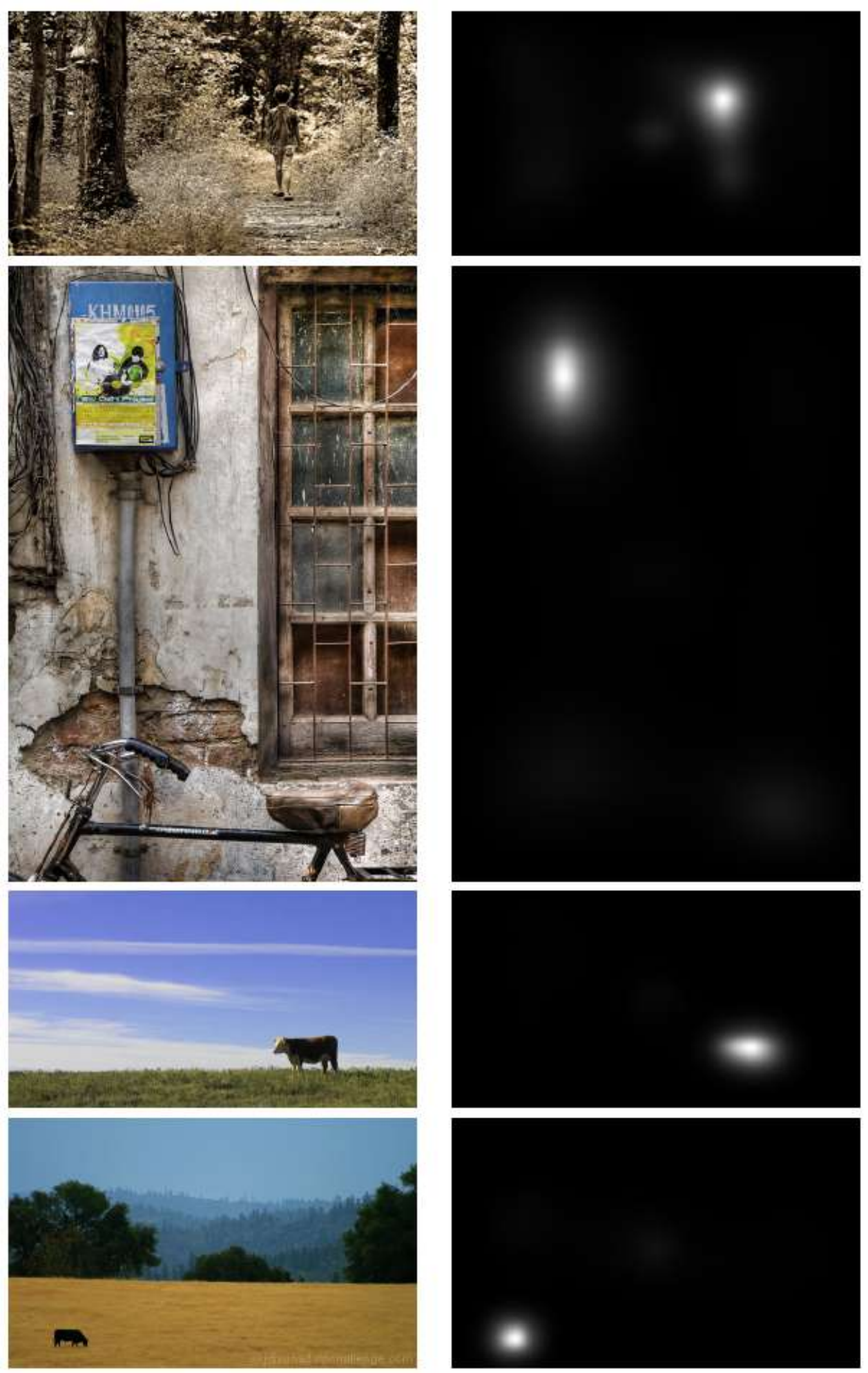}
}
}
&
\resizebox{.518\textwidth}{!}{
\setlength{\fboxrule}{.125 em}
\fbox{
\includegraphics[width=0.5\textwidth]{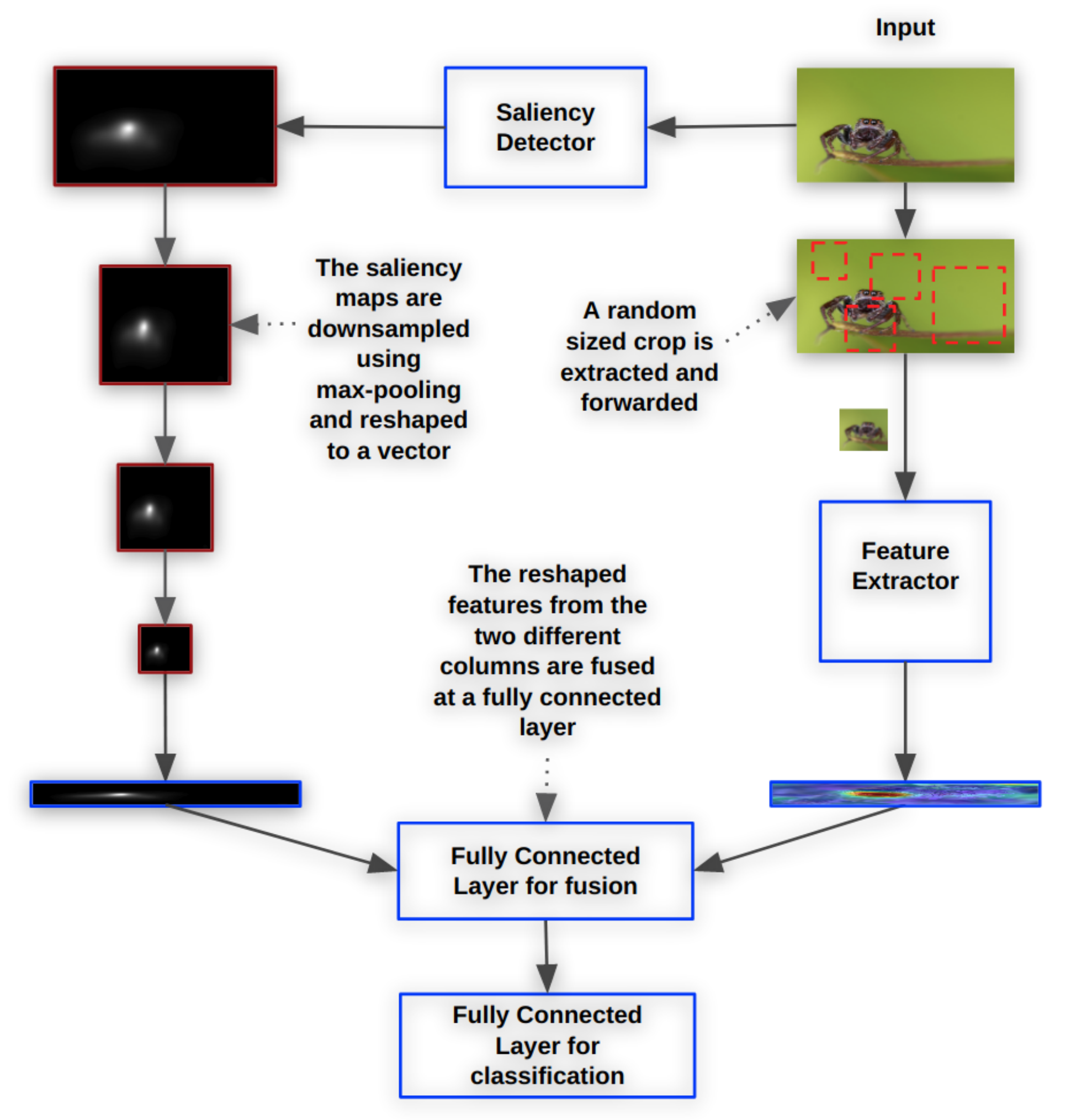}
}
}
\\
(a)&(b)
\end{tabular}
}
\end{center}
\caption{Our Contributions : \textbf{(a)~Input (col 1), saliency maps (col 2) :} Saliency maps are generated  using the method proposed in \cite{cornia2016predicting}.  The position of the main subjects can be obtained from the saliency maps.
\textbf{(b) Our double-column \textsc{CNN} architecture:} One column accepts the regular \textsc{RGB} features and the other column accepts saliency maps. The features from RGB channel are computed using a pre-trained Densenet161~\cite{huang2016densely}, fine-tuned on our datasets. They are fused using a fully-connected layer and finally passed to another final fully-connected layer for classification.}
\label{fig:architecture}
\end{figure}
\noindent The aesthetic quality of a photograph is greatly influenced by its composition, that is a set of styles or attributes which guide the viewer towards the essence of the picture. Analyzing objectively, these styles can be broadly categorized into local or appearance-based~(focus, image-grain,~\etc) and global or geometry-based~(aspect ratio, RoT, framing,~\etc). Figure~\ref{fig:ExampleImages} illustrates some popular styles adopted by photographers for a good composition. \\
In this work, we explore the ability of convolutional neural networks~(CNN) to capture the aesthetic properties of photographic images. Specifically, can CNN based architectures learn both the local or appearance-based~(such as colour) and global or geometry-based~(such as RoT) aspects of photographs and how can we help such architectures capture location specific properties in images?
Motivated by the recent developments in CNNs, our system takes a photograph as an input and predicts its style attributes (ordered by probabilities), as illustrated in Figure~\ref{fig:app}.
There are several applications of automatic photographic style classification. For example, post-processing images and videos, tagging, organizing and mining large collections of photos for artistic, cultural and historical purposes, scene understanding, building assistive-technologies, content creation, cinematography, \etc.

\noindent The traditional approach of using \textsc{CNN}s for natural image classification is to forward a \textit{transformed} version of the input through a series of convolutional, pooling and fully connected layers  and obtain a classification score. The transformation is applied to create a uniform sized input for the network~(crop, warp,~\etc) or to increase variance of the input distribution~(flip, change contrast,~\etc) for better generalization on the test data~\cite{krizhevsky2012imagenet}.
Clearly, such traditional transformations fail to preserve the aesthetic attributes of photographs. For example, a random fixed-sized crop cannot capture the arrangement of subjects within the picture. On the other hand, although warping the input photograph to a fixed size preserves the global context of the subjects better than crop, it distorts the aspect ratio and also smoothens  appearance-based attributes like depth of field or image-grain. 

\noindent This calls for a representation which preserves both the appearance-based and geometry-based properties of a photograph and which generalizes well over test data.
Multiple solutions to these problems have been proposed. In \cite{lu2014rapid}, authors propose a double column \textsc{CNN} architecture, where the first column accepts a cropped patch and the second column accepts a warped version of the entire input. In subsequent work \cite{lu2015deep}, multiple patches are cropped from an input and forwarded through the network. The features from multiple patches are aggregated before the final fully-connected layer for classification. The authors argue that sending multiple patches from the same image encodes more global context than a single random crop. 
More recently, \cite{Ma_2017_CVPR} follow a similar multiple-patch extraction approach, but the patches are selectively extracted based on saliency, pattern-diversity and overlap between the subjects. 
Essentially, these techniques attempt to incorporate global context into the features during a forward pass either by warping the whole input and sending it through an additional column or by providing multiple patches from the input at the same time.  

\noindent Although these traditional double column or multi-patch strategies improve the overall performance, we argue that these networks cannot properly learn the geometry of a photograph.
It is because \textsc{CNN}s, in principle, are designed to be translation invariant~\cite{sabour2017dynamic}. While they can learn how the subjects look like, they cannot capture whether the subjects are rightly positioned. Since the convolutional filters corresponding to a feature map share weights, they become translation-invariant and appearance-dependent. In other words, they are activated for an object irrespective of its location in the image. As a result, they fail to understand photographic attributes like \textsc{RoT}. One option to tackle this could be training a fully-connected network on the full images, but they have too many parameters and are hard to train.  

\noindent Our first contribution in this work is introducing a saliency-based representations~(see Figure~\ref{fig:architecture}(a)) which we call \textbf{Sal-RGB} features. The position or relative geometry of the different subjects in the image are obtained from the saliency maps and then fused with the appearance features coming from a traditional CNN and finally passed to a classifier to identify the overall style of composition of the photograph. By definition, saliency maps are appearance-invariant. On the other hand, by avoiding convolution and fusing them directly with the CNN features we achieve location-cognizance. In Section~\ref{sec:results}, we show that our approach performs better than the SoA in photographic style classification especially for those styles which are geometry-sensitive.

\noindent Our second contribution is a comparative analysis of the traditional approaches for aesthetic categorization of images.  Motivated both from the SoA and recent breakthroughs in deep learning, we implement multiple baselines, by trying different architectures and try to understand and identify the factors that are crucial for encoding the local and global aspects of photographic composition. 

\noindent The rest of the paper is organized as follows. In Section~\ref{sec:reltd}, we summarize the relevant literature in image aesthetic quality prediction. In Section~\ref{sec:arch},  we describe the double column \textsc{CNN} architecture we adopt. In Section~\ref{sec:dataset}, we provide a detailed description of the datasets used. In Section~\ref{sec:results}, we provide  details of the experiments conducted and analyze the results.

\section{Related Work}
\label{sec:reltd}
\noindent Image and video classification has always been a fundamental problem in computer vision. 
Understanding quantifiable visual semantics like the class, position and number of objects in an image were challenging enough and took the majority of focus. However, understanding the subtle, qualitative aspects especially from a creative perspective, due to its even more challenging nature, has only recently started being attacked.

\noindent The initial works in photograph aesthetic assessment relied on explicitly
modelling popular attributes like RoT,
colour harmony, exposure, \etc~\cite{datta2006studying,ke2006design,luo2008photo}. Some recent works address the problem similarly, \textit{i.e} explicitly defining the features but with 
improved performance~\cite{obrador2012towards,dhar2011high,joshi2011aesthetics,san2012leveraging,Karayev2014}. \cite{aydin2015automated} propose a system which predicts the contribution of some photographic attributes towards the overall aesthetic quality of a picture. After estimating the extent of certain compositional attributes, they aggregate the scores for different attributes to predict the overall aesthetic score of a photograph by using a novel calibration technique. 
\cite{murray2012ava} published the Aesthetic Visual Analysis~(\textsc{AVA}) dataset. Improved evaluation due to such datasets and parallel advances in deep learning resulted in a surge of research in this area in the last few years.

\noindent In recent years, deep learning has performed remarkably well in many computer vision tasks like classification~\cite{krizhevsky2012imagenet}, detection~\cite{girshick2015fast}, segmentation~\cite{noh2015learning} and scene understanding~\cite{karpathy2015deep,xu2015show}.  Recent works like~\cite{huang2016densely,he2016deep} have performed well in multi-tasking frameworks for detection and classification. In~\cite{he2016deep}, the authors use a residual framework for tackling training error upon addition of new layers. In ~\cite{huang2016densely} the authors use dense connections by connecting outputs from all the previous layers as input to the next layer.\\
As for many computer vision problems, deep learning has begun to be explored in the domain of image aesthetic assesment as well. Apart from  ~\cite{lu2015deep,lu2014rapid,Ma_2017_CVPR} (discussed in Section~\ref{sec:intro}), in~\cite{kong2016photo}, the authors learn styles and ratings jointly on a new dataset. Their algorithm is based on comparing and ranking a pair of images instead of directly predicting their coarse aesthetic scores. \cite{mai2016composition} propose a network that uses a composition-preserving input mechanism. They introduce an aspect-ratio aware pooling strategy that reshapes each image differently. In~\cite{1707.03981}, the authors propose a network that predicts the overall aesthetic score and eight style attributes, jointly. Additionally, they use gradient-based feature visualization techniques to understand the correlation of different attributes with image locations.

\noindent In principle, our pipeline is similar to \cite{lu2015deep,lu2014rapid,Karayev2014} in the sense that we also perform a neural style prediction on the AVA dataset. However, our work differs in two important aspects. First, in the overall style prediction, our Sal-RGB features perform better than the strategies that use generic features~\cite{Karayev2014}, the double column~\cite{lu2014rapid} or multi-patch aggregation~\cite{lu2015deep}. Second, unlike ~\cite{lu2014rapid, lu2015deep} we analyze individual attributes and evaluate our strategy on multiple datasets. 
\section{Network Architecture }
\label{sec:arch}
\noindent In this section, we describe our architecture, as illustrated in Figure~\ref{fig:architecture}(b).
\noindent Our architecture consists of three main blocks --- the saliency detector, the double-column feature-extractor
and the classifier.
\subsection{Saliency Detector}
\noindent We compute the saliency maps using the method proposed in \cite{cornia2016predicting}. Motivated from recent attention based models \cite{xu2015show} that processes some regions of the input more attentively than others, the authors propose a CNN-LSTM (long and short term memory network) framework for saliency detection. LSTMs are applied to sequential inputs where output from previous states are combined with inputs to the next state using dot products. In this work, the authors modify the standard LSTM such that they accept a sequence of spatial data (patches extracted from different locations in the image) and combine them using convolutions instead of dot products. Additionally, they introduce a center-prior component, that handles the tendency of humans to fix attention at the center region of an image.  Some outputs from the system can be found in Figure~\ref{fig:architecture}(a), second column. 
\subsection{Feature Extractor}
The feature extractor consists of two parallel and independent columns, one for the saliency map and the other for raw RGB input.\\
\noindent \textbf{Saliency Column : } The saliency column consists of two max-pooling layers that downsample the input from $224 \times 224$ to $56 \times 56$ as shown in ~\ref{fig:architecture}(b). Instead of max-pooling, we tried strided convolutions as they are known to capture low level details better than pooling~\cite{johnson2016perceptual}. But pooling gave better results in our case which perhaps indicates that the salient position was more important than the level of detail captured. \\
\noindent \textbf {RGB Column : }
We choose the DenseNet161 \cite{huang2016densely} network for its superior performance in the ImageNet challenge. Very deep networks suffer from the \textit{vanishing-gradient} problem 
\textit{i.e.} gradual loss of information as the input passes through several intermediate layers. Recent works like~\cite{he2016deep,srivastava2015training} address this problem by explicitly passing information between layers or by dropping random layers while training. The DenseNet is different from the traditional \textsc{CNN}s in the manner in which each layer receives input from the previous layers. The $l^{th}$ layer in DenseNet receives as input, the concatenated output from all previous $l-1$ layers. \\
\noindent We replace the last fully-connected layer from DenseNet with our classifier described in Section~\ref{sec:classifier} and use the remaining as a feature extractor.
Since we have less training images, we fine-tune a model pre-trained on ImageNet on our dataset instead of training from scratch. This works since the lower level features like edges and corners are generic image features and can be used for aesthetic tasks too. 
\subsection{Classifier}
\label{sec:classifier}
\noindent Feature-maps from the two columns are concatenated and fused together using a fully-connected layer. A second and final fully-connected layer is used as a classifier. During training, we use the standard cross-entropy loss function and the gradient is back-propagated to the two columns.
\section{Datasets} 
\label{sec:dataset}
\noindent We use two standard datasets for evaluation --- AVA Style and Flickr Style.
\noindent AVA~\cite{murray2012ava} is a dataset containing $250,000$ photographs, selected from www.dpchallenge.com. Dpchallenge is a forum for photographers. Users rate each photograph during the challenge on a scale of $10$ and post feedback during and after the challenge. 

\noindent Of these $250,000$ photographs, the authors manually select $72$ challenges, corresponding to $14$ different photographic styles as illustrated in Figure~\ref{fig:ExampleImages} and create a subset called AVA Style containing about $14,000$ images. 
While training images in the subset are annotated with a single label, the test images have multiple labels associated with them making them unsuitable for popular evaluation frameworks used for single-label multi-class classifiers. 

 \noindent Flickr Style~\cite{Karayev2014} is a collection of $80,000$ images of $20$ visual styles. The styles span across multiple concepts such as optical techniques~(Macro, Bokeh,~\etc), atmosphere~(Hazy, Sunny,~\etc), mood~(Serene, Melancholy,~\etc), composition styles~(Minimal, Geometric,~\etc), colour~(Pastel, Bright,~\etc) and genre~(Noir, Romantic,~\etc). Flickr Style is a more complex dataset than AVA not only because it has more classes, but because some of the classes like Horror, Romantic and Serene are subjective concepts and difficult to encode objectively.
\section{Experiments} \label{sec:results}
We investigate two different aspects of the problem. First, in Section~\ref{sec:ResStyleClass} we report the overall performance of our features using mean average precision~(MAP). Second, in Section~\ref{sec:classwisepre} we observe the per-class precision~(PCP) scores to understand how our features affect individual photographic attributes. 
\noindent For comparison, we use MAP reported in \cite{Karayev2014,lu2014rapid,lu2015deep}. PCP is compared only with \cite{Karayev2014} since the implementations were unavailable for \cite{lu2014rapid,lu2015deep}. 
Additionally, we implement the following two benchmarks to evaluate our approach. 
\begin{itemize}[noitemsep,topsep=0pt,parsep=0pt,partopsep=0pt,leftmargin=*]
    \item \textbf{DenseNet161, ResNet152} : These are off-the-shelf implementations~\cite{huang2016densely, he2016deep} finetuned on our dataset and takes only RGB representation as input. These were chosen since they achieve the least error rates for ImageNet classification. 
    \item \textbf{RAPID++} : Following~\cite{lu2014rapid}, we implemented a two-column network. Each column takes as input, random crops and the whole image, as local and global representations, respectively. But, we used DenseNet161 architecture for the two columns whereas in the original work the authors use a shallower architecture with only three layers. We choose this as a benchmark in order to observe how their algorithm performs with a deeper architecture. 
\end{itemize}
\noindent We train style classifiers on the AVA Style and Flickr Style datasets. 
The train-test partitions are followed from the original papers \cite{murray2012ava, Karayev2014}. \\
\noindent For AVA, We use $11270$ images for training and validation and $2573$ images for testing. For Flickr Style we use $64000$ images for training and $16000$ images for testing. 
For testing, we follow the approach adopted by ~\cite{lu2014rapid,lu2015deep}. $50$ patches are extracted from the test-image and each patch is passed through the network.
The results are averaged to achieve the final scores.
\subsection{Style Classification}
\label{sec:ResStyleClass}
The scores are reported in terms of Mean Average Precision (MAP). MAP refers to the average of per-class precision. 
The results are reported in~Table~\ref{tab:allNetStyle}. We observe that our method outperforms the SoA \cite{Karayev2014,lu2014rapid,lu2015deep} significantly. But, our own baselines perform more or less 
\begin{wraptable}[14 ]{l}{0.5\textwidth}
\vspace{-5 pt}
\caption{\textbf{Style Classification : Comparison with the SoA :} The results are reported in terms of Mean Average Precision(average of per class precision). We observe that for both the datasets, our method performs better than the state of the art. Flickr Style was not used in \cite{lu2014rapid,lu2015deep}.}
\def\arraystretch{1.2}
\begin{center}
\resizebox{.5\textwidth}{!}{
\begin{tabular}{|c|c|c|c|}
\hline
\textbf{Network} & \textbf{Augmentation} & \textbf{AVA} & \textbf{Flickr Style}\\
\hline\hline
Fusion~\cite{Karayev2014} & centre crop & 58.10 & 36.80 \\
RAPID~\cite{lu2014rapid} & random crop, warp &  56.81 & -\\
Multi-Patch~\cite{lu2015deep} & random crop & 64.07 & - \\
\hline
 DenseNet161~\cite{huang2016densely} & random crop & 71.68 & 43.83\\
 ResNet152~\cite{he2016deep} & random crop & 70.57 & 43.65\\
 RAPID++ & random crop, warp & 70.48 & 41.93 \\
 \hline
 \textbf{Sal-RGB}  & random crop & 71.82 &  43.45 \\
\hline
\end{tabular}
}
\end{center}
\vspace{-20 pt}
\label{tab:allNetStyle}
\end{wraptable}
equally well. We deduce that for the improvement of MAP, the maximum impact is made by a more sophisticated CNN, followed by the location specific saliency. Both ResNet~\cite{he2016deep} and DenseNet~\cite{huang2016densely} are residual networks and learn complex representations due to their very deep architectures.  Such representations are crucial for learning photographic attributes, which have many overlapping properties~(less inter-class variance). \\
\noindent From these results,  one might argue that the improvement can be attributed largely to a better CNN, and so what does Sal-RGB bring to the representation ? We address this issue in Section~\ref{sec:classwisepre}. 
\subsection{Per-class Precision Scores }
\label{sec:classwisepre}
\noindent In ~\cite{Karayev2014}, the authors report per-class precision (PCP) scores on \textsc{AVA} Style and Flickr Style. We compare our algorithm with those results in Table \ref{tab:per_class_flickr}. \\
\noindent We observe that our method outperforms \cite{Karayev2014} in almost all categories on both datasets. For the AVA Style dataset, a significant improvement is observed in the appearance-based categories like complementary colours, duotones, image grain, \etc.  Yet again, our own baselines DenseNet, ResNet and RAPID++ perform equally well in most categories except for RoT. For this category, Sal-RGB outperforms all others by a significant margin. This is an important result, since unlike others, RoT is a purely geometric attribute and important for image aesthetics and photography. A significant improvement in this category is a confirmation of our claim that the proposed approach efficiently encodes the geometry of a photograph. We highlight these observations in the bar plot beside Table ~\ref{tab:per_class_flickr}.
\begin{table}
\caption{\textbf{PCP for AVA Style} : Sal-RGB outperforms the SoA~\cite{Karayev2014} by a significant margin in every category. Our own baselines DenseNet~\cite{huang2016densely}, ResNet~\cite{he2016deep}, RAPID++ perform equally well for almost all categories except RoT, for which Sal-RGB performs much better. The bar plot on the right shows the relative improvements in overall MAP and RoT respectively. }
\def\arraystretch{1.2}
\begin{center}
\resizebox{.95\textwidth}{!}{
\begin{tabular}{|c||p{0.1\textwidth}||p{0.1\textwidth} p{0.1\textwidth} p{0.1\textwidth}||p{0.1\textwidth}|}
\hline 
Styles & Fusion(SoA) & Densenet161   & ResNet152  &RAPID++ & \textbf{Sal-RGB}  \tabularnewline
\hline 
\hline 
Complementary\_Colors & \textcolor{black}{46.90} & 62.33 & 62.15 & 61.49 & 61.41\tabularnewline
\hline 
Duotones & \textcolor{black}{67.60} & 86.58 & 84.82 & 84.77 & 87.58\tabularnewline
\hline 
HDR & \textcolor{black}{66.90} & 74.95 & 70.08 & 71.51 & 72.86\tabularnewline
\hline 
Image\_Grain & \textcolor{black}{64.70} & 81.55 & 79.48 & 83.15 & 82.20\tabularnewline
\hline 
Light\_On\_White & \textcolor{black}{90.80} & 84.69 & 83.41 & 85.64 & 82.99\tabularnewline
\hline 
Long\_Exposure & \textcolor{black}{45.30} & 64.16 & 65.38 & 63.94 & 61.94\tabularnewline
\hline 
Macro & \textcolor{black}{47.80} & 64.89 & 65.52 & 64.90 & 66.58\tabularnewline
\hline 
Motion\_Blur & \textcolor{black}{47.80} & 63.93 & 62.12 & 61.21 & 61.98\tabularnewline
\hline 
Negative\_Image & \textcolor{black}{59.50} & 87.40 & 86.11 & 82.01 & 87.71\tabularnewline
\hline 
\textbf{Rule\_of\_Thirds} & \textbf{\textcolor{black}{35.20}} & \textbf{33.16} & \textbf{34.27} & \textbf{34.02} & \textbf{41.68}\tabularnewline
\hline 
Shallow\_DOF & \textcolor{black}{62.40} & 82.08 & 82.42 & 82.95 & 82.39\tabularnewline
\hline 
Silhouettes & \textcolor{black}{79.10} & 93.73 & 92.49 & 91.14 & 93.05\tabularnewline
\hline 
Soft\_Focus & \textcolor{black}{31.20} & 49.89 & 44.91 & 44.57 & 46.41\tabularnewline
\hline 
Vanishing\_Point & \textcolor{black}{68.40} & 74.16 & 74.80 & 75.45 & 76.76\tabularnewline
\hline 
\end{tabular}
\vspace{-20 pt}
\quad
\begin{tabular}{c}
     \includegraphics[width = .7\textwidth]{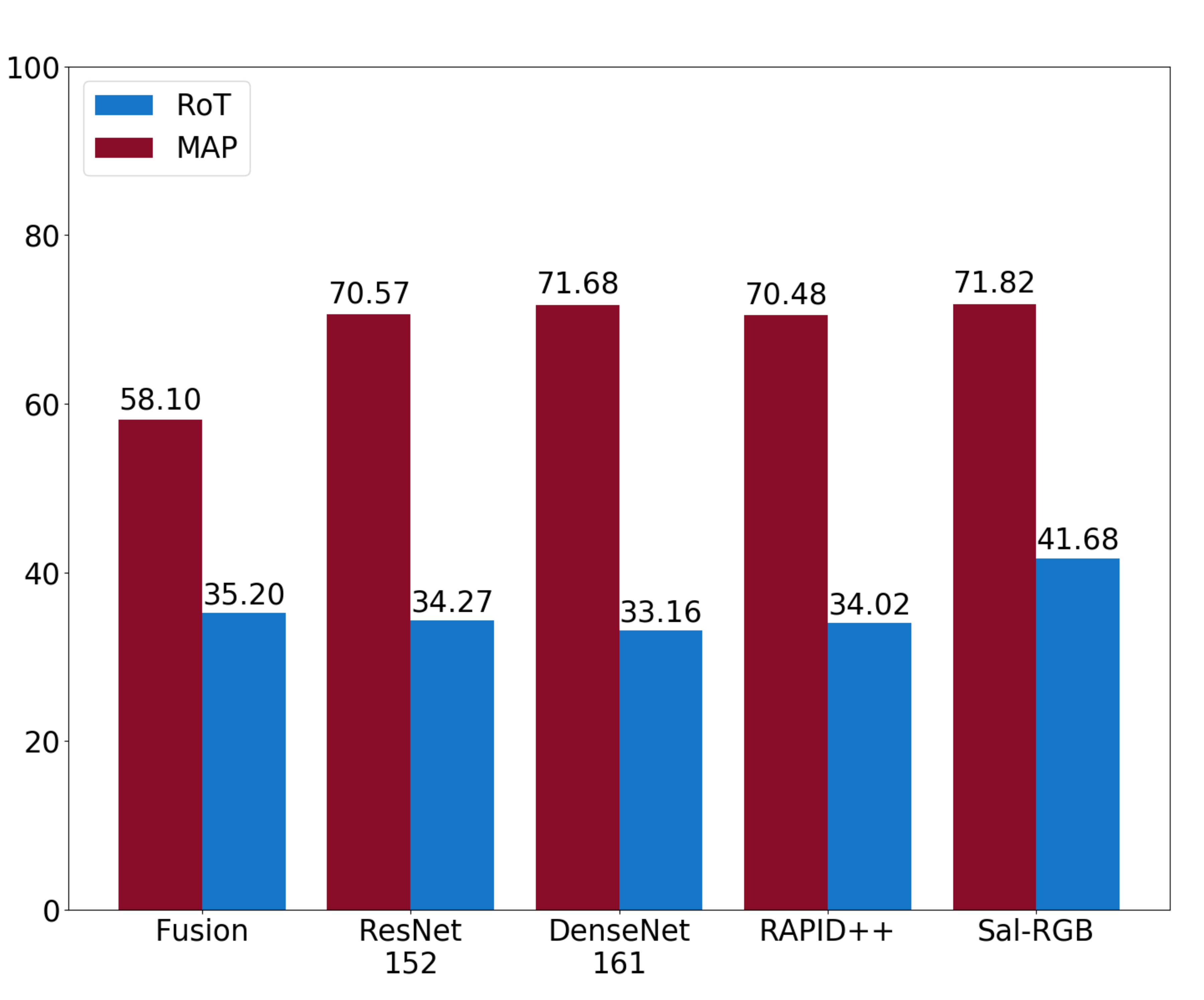}
\end{tabular}
}
\end{center}
\label{tab:per_class_flickr}
\end{table}
\subsection{Limitations}
 \label{sec:limitations}
 \noindent We tried to understand the limitations of our approach by plotting the confusion matrix for the different attributes of AVA.\\
   \begin{wrapfigure}[18]{r}{.45\textwidth}
   \vspace{-20 pt}
      \centering
      \resizebox{.45\textwidth}{!}{
      \includegraphics{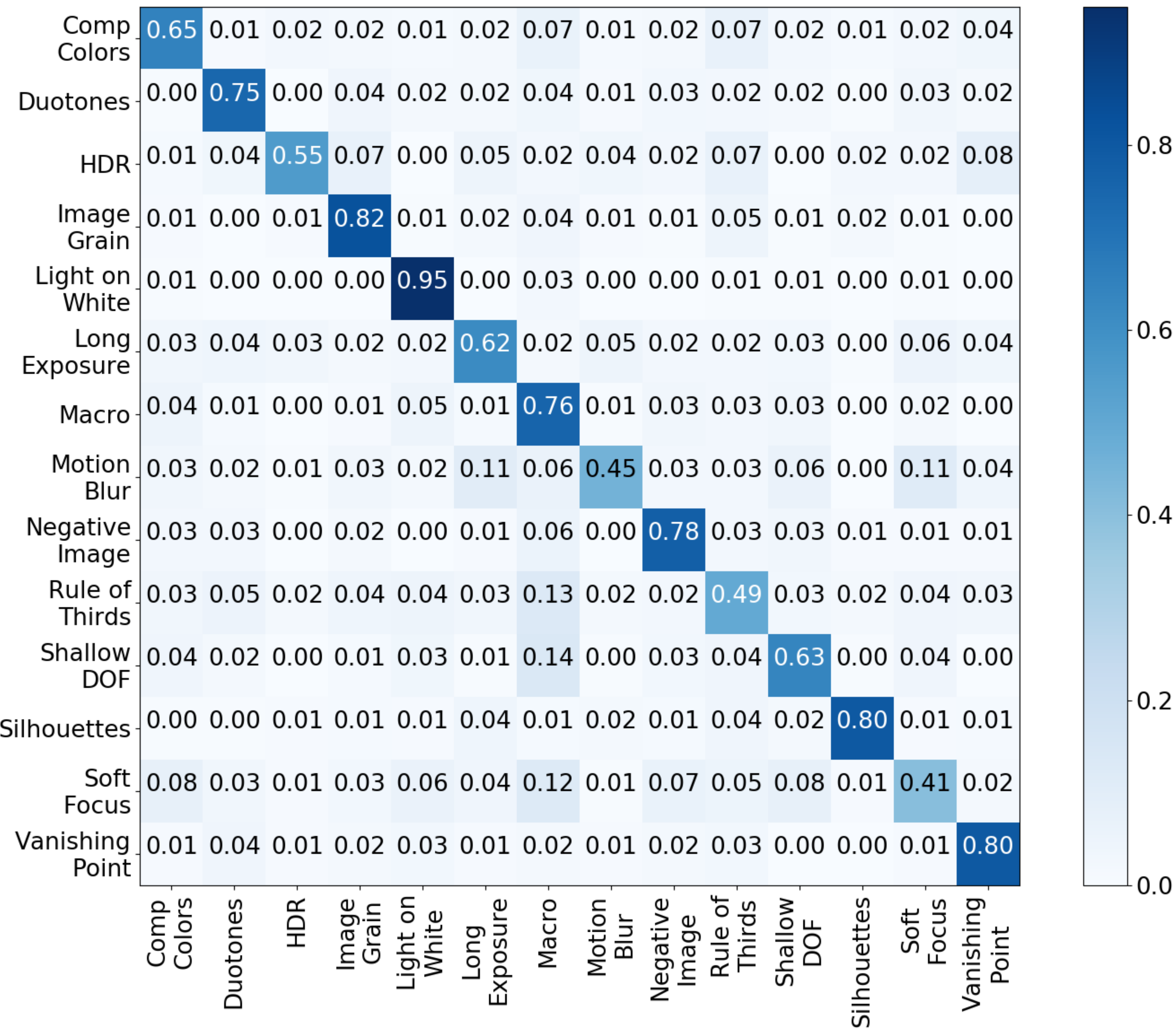}
      }
      \vspace{-20 pt}
      \caption{\textbf{Confusion matrix for AVA Style with our model:} For a test sample, the rows correspond to the real class and the columns correspond to the predicted class. The values are computed over 2573 test samples of AVA and then normalized.} 
      \label{fig:conf_mat}
  \end{wrapfigure}
  \begin{itemize}[noitemsep,topsep=0pt,parsep=0pt,partopsep=0pt,leftmargin=*]
  \item The strongest classes are Light on White, Silhouettes, Vanishing Points.
  The weakest are Motion Blur and Soft Focus. 
  \item Long Exposure and Motion Blur get confused with each other, which
  makes sense, since both attributes are captured using a slow
  shutter speed and mostly at night.
  \item Shallow DOF, Soft Focus and Macro are mutually confused classes, which is justified as all of them involve blur.
  \item The poorly performing classes have a high false-positive rate. We blame this on two factors. First, some classes such as Motion Blur and Soft-Focus have less samples as compared to others. Secondly, we observe that there is some ambiguity in the annotation of the training data of ~\textsc{AVA}. They are associated with a single label. But usually, most of the good photographs are captured with an interplay between multiple attributes.  For example a macro image could very well conform to RoT or depth of field. Thus a single annotation incorporates undesired penalties to the loss during training the network and creates confusions during prediction.
  \end{itemize}
\section{Conclusion}
There are many potential applications of an automatic style and aesthetic quality estimator in the domain of digital photography such as interactive cameras, automated photo correction \etc.
Our system can be directly extended to video-processing for predicting shot-styles. For example, Figures~\ref{fig:app} illustrates the aesthetic analysis of a shot taken from Majid Majidi's movie Colours of Paradise.
\noindent As future work, there are many possible directions. Generalizing the model to more style attributes could be one. Extending the system to the domain of video and 360 images would also be possible. A thorough mathematical analysis of seemingly intangible and subjective concepts in art and subsequently fixing ambiguities in the data-annotation could be another. We hope that this area will become more active in the future with its challenging and interesting set of problems.\footnote{This publication has emanated from research conducted with the financial support of Science Foundation Ireland (SFI) under the Grant Number 15/RP/2776}
\bibliographystyle{apalike}
{\footnotesize
\bibliography{ms}}

\end{document}